\documentclass[conference]{IEEEtran}
\usepackage{blindtext, graphicx}
\usepackage{amsmath}

\ifCLASSINFOpdf

\fi

\begin{document}
%
\title{Medical image denoising using convolutional denoising autoencoders}

\author{\IEEEauthorblockN{Lovedeep Gondara}
\IEEEauthorblockA{Department of Computer Science\\
Simon Fraser University\\
lgondara@sfu.ca}}

\maketitle

\begin{abstract}
Image denoising is an important pre-processing step in medical image analysis. Different algorithms have been proposed in past three decades with varying denoising performances. More recently, having outperformed all conventional methods, deep learning based models have shown a great promise. These methods are however limited for requirement of large training sample size and high computational costs. In this paper we show that using small sample size, denoising autoencoders constructed using convolutional layers can be used for efficient denoising of medical images. Heterogeneous images can be combined to boost sample size for increased denoising performance. Simplest of networks can reconstruct images with corruption levels so high that noise and signal are not differentiable to human eye.

\end{abstract}


%
\IEEEpeerreviewmaketitle

\section{Introduction}

Medical imaging including X-rays, Magnetic Resonance Imaging (MRI), Computer Tomography (CT), ultrasound etc. are susceptible to noise \cite{sanches2008}. Reasons vary from use of different image acquisition techniques to  attempts at decreasing patients exposure to radiation. As the amount of radiation is decreased, noise increases \cite{Agostinelli2013}. Denoising is often required for proper image analysis, both by humans and machines. 

Image denoising, being a classical problem in computer vision has been studied in detail. Various methods exist, ranging from models based on partial differential equations (PDEs) \cite{peronaPDE1990, rudinPDE1994,subakanPDE2007}, domain transformations such as wavelets \cite{donoho_wavelet1995}, DCT \cite{yaroslavsky2001}, BLS-GSM \cite{portialla2003} etc., non local techniques including NL-means \cite{zhang2002, buades2005}, combination of non local means and domain transformations such as BM3D \cite{dabov_bm3d2007} and a family of models exploiting sparse coding techniques \cite{sparse_coding1997,sparse_coding2006,sparse_coding2009}. All methods share a common goal, expressed as

\begin{equation}
z=x+y    
\end{equation}

Where $z$ is the noisy image produced as a sum of original image $x$ and some noise $y$. Most methods try to approximate $x$ using $z$ as close as possible. IN most cases, $y$ is assumed to be generated from a well defined process.

With recent developments in deep learning \cite{deeplearning1_alex2012,deeplearning2_hinton2012,deeplearning3_sutskever2014,deeplearning4_bengio2007,deeplearning5_glorot2011}, results from models based on deep architectures have been promising. Autoencoders have been used for image denoising \cite{vincent_auto2008,vincent_auto2010,xier_denoise_auto2012,cho_denoise_auto2013}. They easily outperform conventional denoising methods and are less restrictive for specification of noise generative processes. Denoising autoencoders constructed using convolutional layers have better image denoising performance for their ability to exploit strong spatial correlations. 

In this paper we present empirical evidence that stacked denoising autoencoders built using convolutional layers work well for small sample sizes, typical of medical image databases. Which is in contrary to the belief that for optimal performance, very large training datasets are needed for models based on deep architectures. We also show that these methods can recover signal even when noise levels are very high, at the point where most other denoising methods would fail. 

Rest of this paper is organized as following, next section discusses related work in image denoising using deep architectures. Section III introduces autoencoders and their variants. Section IV explains our experimental set-up and details our empirical evaluation and section V presents our conclusions and directions for future work.

\section{Related work}

 Although BM3D \cite{dabov_bm3d2007} is considered state-of-the-art in image denoising and is a very well engineered method, Burger et al. \cite{burger_denoise2012} showed that a plain multi layer perceptron (MLP) can achieve similar denoising performance.
 
 Denoising autoencoders are a recent addition to image denoising literature. Used as a building block for deep networks, they were introduced by Vincent et al. \cite{vincent_auto2008} as an extension to classic autoencoders. It was shown that denoising autoencoders can be stacked  \cite{vincent_auto2010} to form a deep network by feeding the output of one denoising autoencoder to the one below it. 
 
 Jain et al. \cite{jain_denoise2009} proposed image denoising using convolutional neural networks. It was observed that using a small sample of training images, performance at par or better than state-of-the-art based on wavelets and Markov random fields can be achieved. Xie et al. \cite{xier_denoise_auto2012} used stacked sparse autoencoders for image denoising and inpainting, it performed at par with K-SVD. Agostenelli et al. \cite{Agostinelli2013}  experimented with adaptive multi column deep neural networks for image denoising, built using combination of stacked sparse autoencoders. This system was shown to be robust for different noise types. 
 
\section{Preliminaries}

\subsection{Autoencoders}
An autoencoder is a type of neural network that tries to learn an approximation to identity function using backpropagation, i.e. given a set of unlabeled training inputs ${x^{(1)},x^{(2)},...,x^{(n)}}$, it uses

\begin{equation}
    z^{(i)}=x^{(i)}
\end{equation}

An autoencoder first takes an input $x \in [0,1]^d$ and maps(encode) it to a hidden representation $y \in [0,1]^{d'}$ using deterministic mapping, such as

\begin{equation}
    y=s(Wx+b)
\end{equation}

where $s$ can be any non linear function. Latent representation $y$ is then mapped back(decode) into a reconstruction $z$, which is of same shape as $x$ using similar mapping.

\begin{equation} \label{primeq}
    z=s(W'y+b')
\end{equation}

In \eqref{primeq}, prime symbol is not a matrix transpose. Model parameters ($W,W',b,b'$) are optimized to minimize reconstruction error, which can be assessed using different loss functions such as squared error or cross-entropy.

\begin{figure}[h]
  \centering
  \includegraphics[scale=.3]{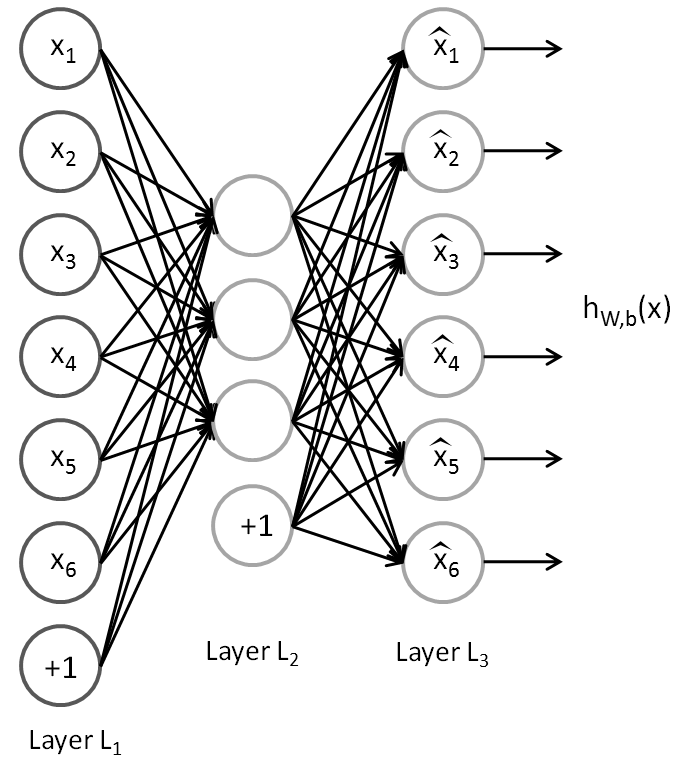}
  \caption{A basic autoencoder}
   \label{autoencoder_fig}
\end{figure}

Basic architecture of an autoencoder is shown in Fig. \ref{autoencoder_fig} \cite{autoencoder_fig}

Here layer $L_1$ is input layer which is encoded in layer $L_2$ using latent representation and input is reconstructed at $L_3$.

Using number of hidden units lower than inputs forces autoencoder to learn a compressed approximation. Mostly an autoencoder learns low dimensional representation very similar to Principal Component Analysis (PCA). Having hidden units larger than number of inputs can still discover useful insights by imposing certain sparsity constraints. 

\subsubsection{Denoising Autoencoders}

Denoising autoencoder is a stochastic extension to classic autoencoder \cite{vincent_auto2008}, that is we force the model to learn reconstruction of input given its noisy version. A stochastic corruption process randomly sets some of the inputs to zero, forcing denoising autoencoder to predict missing(corrupted) values for randomly selected subsets of missing patterns. 

\begin{figure}[h]
  \centering
  \includegraphics[scale=.5]{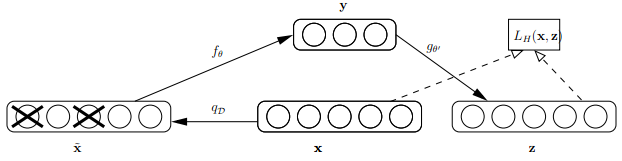}
  \caption{Denoising autoencoder, some inputs are set to missing}
    \label{denautoencoder_fig}
\end{figure}

Basic architecture of a denoising autoencoder is shown in Fig. \ref{denautoencoder_fig}

\begin{figure}[h]
  \centering
  \includegraphics[scale=.5]{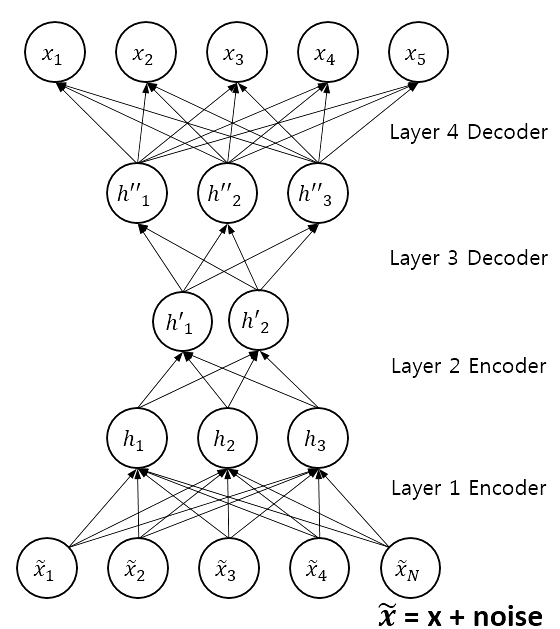}
  \caption{A stacked denoising autoencoder}
   \label{sautoencoder_fig}
\end{figure}

Denoising autoencoders can be stacked to create  a deep network (stacked denoising autoencoder) \cite{vincent_auto2010} shown in Fig. \ref{sautoencoder_fig} \cite{sautoencoder_fig}.

Output from the layer below is fed to the current layer and training is done layer wise.

\subsubsection{Convolutional autoencoder}
Convolutional autoencoders \cite{masci2011} are based on standard autoencoder architecture with convolutional \emph{encoding} and \emph{decoding} layers. Compared to classic autoencoders, convolutional autoencoders are better suited for image processing as they utilize full capability of convolutional neural networks to exploit image structure.

In convolutional autoencoders, weights are shared among all input locations which helps preserve local spatiality. Representation of $i$th feature map is given as 

\begin{equation}
    h^i=s(x * W^i+b^i).
\end{equation}

where bias is broadcasted to whole map, $*$ denotes convolution (2D) and $s$ is an activation. Single bias per latent map is used and reconstruction is obtained as 

\begin{equation}
    y=s(\sum_{i \in H} h^i * \tilde{W}^i+c)
\end{equation}

where $c$ is bias per input channel, $H$ is group of latent feature maps, $\tilde{W}$ is flip operation over both weight dimensions. 

Backpropogation is used for computation of gradient of the error function with respect to the parameters. 

\section{Evaluation}

\subsection{Data}
\begin{figure}[h]
  \centering
  \includegraphics[scale=.5]{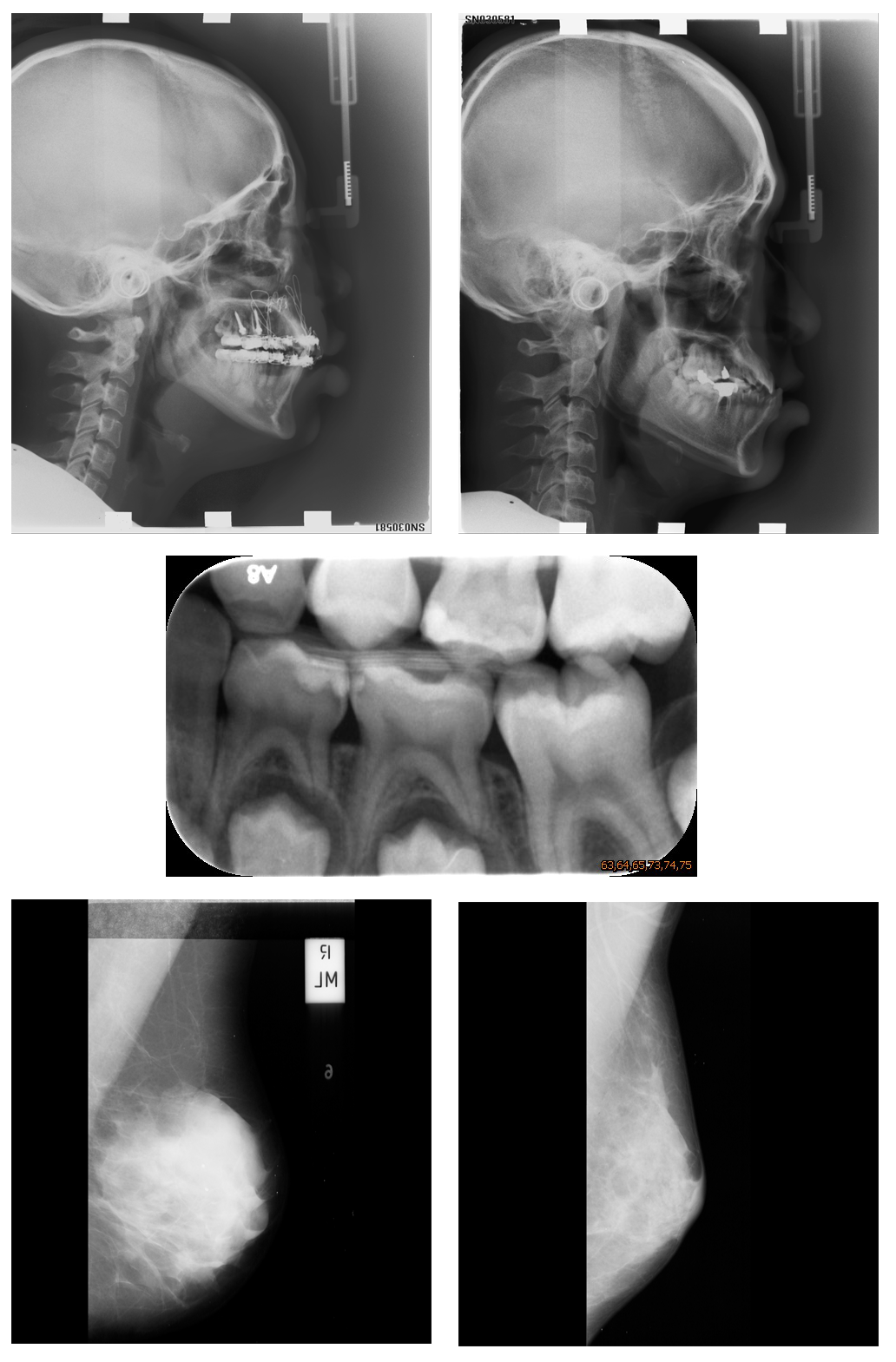}
  \caption{Random sample of medical images from datasets MMM and DX, rows 1 and 2 show X-ray images from DX, whereas row 3 shows mammograms from MMM}
\label{random_real}
\end{figure}

We used two datasets, mini-MIAS database of mammograms(MMM) \cite{minimias} and a dental radiography database(DX) \cite{wangxray}. MMM has 322 images of 1024 $\times$ 1024 resolution and DX has 400 cephalometric X-ray images collected from 400 patients with a resolution of 1935 $\times$ 2400. Random images from both datasets are shown in Fig. \ref{random_real}.

\subsection{Experimental setup}
All images were processed prior to modelling. Pre-processing consisted of resizing all images to 64 $\times$ 64 for computational resource reasons. Different parameters detailed in Table \ref{datasets} were used for corruption.

\begin{table}[h]
\centering
\caption{Dataset perturbations}
\label{datasets}
\begin{tabular}{|l|l|}
\hline
Noise type & corruption parameters        \\ \hline
Gaussian   & $p$=0.1, $\mu=0$, $\sigma=1$ \\ \hline
Gaussian   & $p$=0.5, $\mu=0$, $\sigma=1$ \\ \hline
Gaussian   & $p$=0.2, $\mu=0$, $\sigma=2$ \\ \hline
Gaussian   & $p$=0.2, $\mu=0$, $\sigma=5$ \\ \hline
Poisson    & $p$=0.2, $\lambda=1$         \\ \hline
Poisson    & $p$=0.2, $\lambda=5$         \\ \hline
\end{tabular}
\vspace{5mm}
\\$p$ is proportion of noise introduced, $\sigma$ and $\mu$ are standard deviation and mean of normal distribution and $\lambda$ is mean of Poisson distribution

\end{table}

\begin{figure}[h]
  \centering
  \includegraphics[scale=.5]{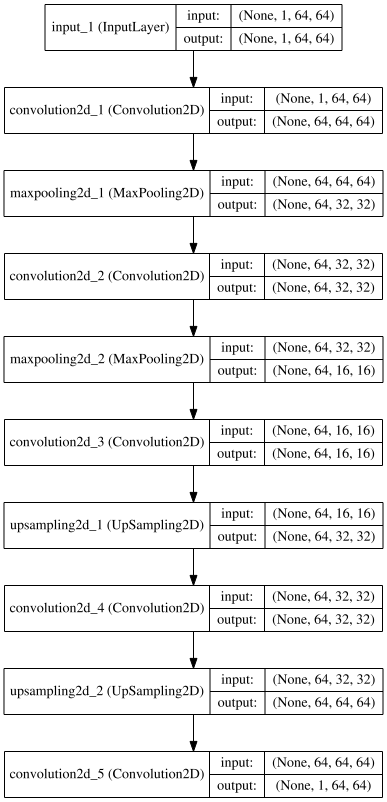}
  \caption{Architecture of CNN DAE used}
  \label{model_arch}
\end{figure}

Instead of corrupting a single image at a time, flattened dataset with each row representing an image was corrupted, hence simultaneously perturbing all images. Corrupted datasets were then used for modelling. Relatively simple architecture was used for convolutional denoising autoencoder (CNN DAE), shown in Fig. \ref{model_arch}. 

Keras \cite{keras} was used for implementing this model on an Acer Aspire M5 notebook (Intel Core i5-4200U, 10 GB RAM, no GPU). Images were compared using structural similarity index measure(SSIM) instead of peak signal to noise ratio (PSNR) for its consistency and accuracy \cite{ssim}. A composite index of three measures, SSIM estimates the visual effects of shifts in image luminance, contrast and other remaining errors, collectively called structural changes. For original and coded signals $x$ and $y$, SSIM is given as

\begin{equation}
    SSIM(x,y)=[l(x,y)]^\alpha[c(x,y)]^\beta[s(x,y)]^\gamma
\end{equation}

where $\alpha, \beta$ and $\gamma > 0 $ control the relative significance of each of three terms in SSIM and $l$, $c$ and $s$ are luminance, contrast and structural components calculated as 

\begin{equation}
    l(x,y) = \dfrac{2\mu_x \mu_y +C_1}{\mu^2_x+\mu^2_y+C_1}
\end{equation}

\begin{equation}
    c(x,y) = \dfrac{2\sigma_x \sigma_y +C_2}{\sigma^2_x+\sigma^2_y+C_2}
\end{equation}

\begin{equation}
    s(x,y) = \dfrac{2\sigma_{xy}+C_3}{\sigma_x\sigma_y+C_3}
\end{equation}

where $\mu_x$ and $\mu_y$ represents the mean of original and coded image, $\sigma_x$ and $\sigma_y$ are standard deviation and $\sigma_{xy}$ is the covariance of two images.

Basic settings were kept constant with 100 epochs and a batch size of 10. No fine-tuning was performed to get comparison results on a basic architecture, that should be easy to implement even by a naive user. Mean of SSIM scores over the set of test images is reported for comparison.

\subsection{Empirical evaluation}
For baseline comparison, images corrupted with lowest noise level ($\mu=0, \sigma=1, p=0.1$) were used. To keep similar sample size for training, we used 300 images from each of the datasets, leaving us with 22 for testing in MMM and 100 in DX. 

\begin{figure}[h]
  \centering
  \includegraphics[scale=0.8]{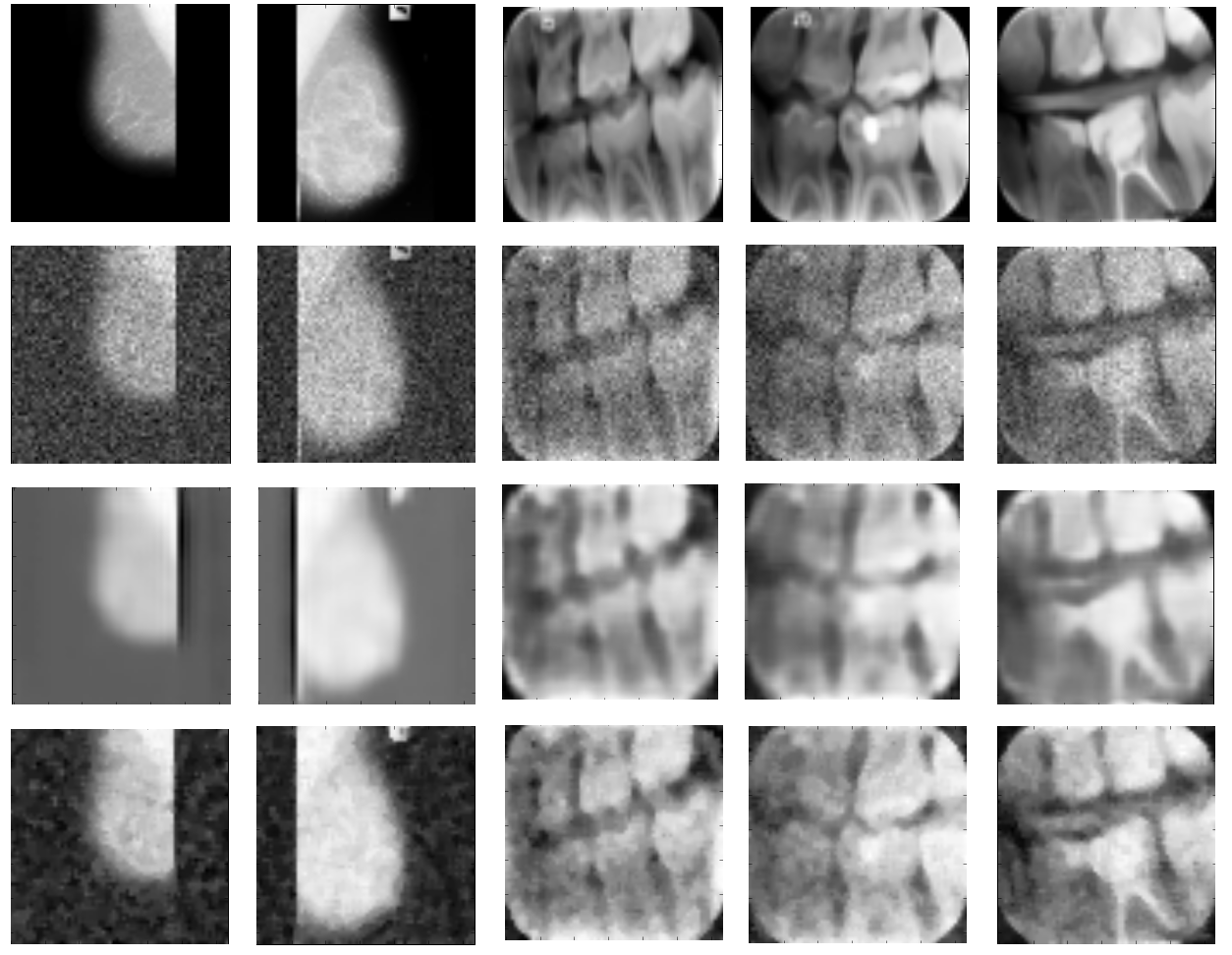}
  \caption{Denoising results on both datasets, top row shows real images with second row showing the noisier version ($\mu=0, \sigma=1, p=0.1$), third row shows images denoised using CNN DAE and fourth row shows results of applying a median filter}
  \label{smalldata}
\end{figure}

Using a batch size of 10 and 100 epochs, denoising results are presented in Fig. \ref{smalldata} and Table \ref{smalldatatable}.

\begin{table}[h]
\centering
\caption{Mean SSIM scores for test images from MMM and DX datasets}
\label{smalldatatable}
\begin{tabular}{|l|l|l|}
\hline
Image type    & MMM & DX \\ \hline
Noisy         & 0.45       & 0.62   \\ \hline
CNN DAE       & 0.81       & 0.88   \\ \hline
Median filter & 0.73       & 0.86  \\
\hline
\end{tabular}
\end{table}

\begin{figure}[h]
  \centering
  \includegraphics[scale=0.5]{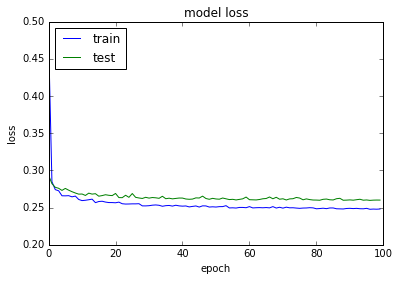}
  \caption{Training and validation loss from 100 epochs using a batchsize of 10}
  \label{loss1}
\end{figure}

Results show an increased denoising performance using this simple architecture on small datasets over the use of median filter, which is most often used for this type of noise.

Model converged nicely for the given noise levels and sample size, shown in Fig. \ref{loss1}. It can bee seen that even using 50 epochs, reducing training time in half, we would have got similar results.

To test if increased sample size by combining heterogeneous data sources would have an impact on denoising performance, we combined both datasets with 721 images for training and 100 for testing.

\begin{figure}[h]
  \centering
  \includegraphics[scale=1]{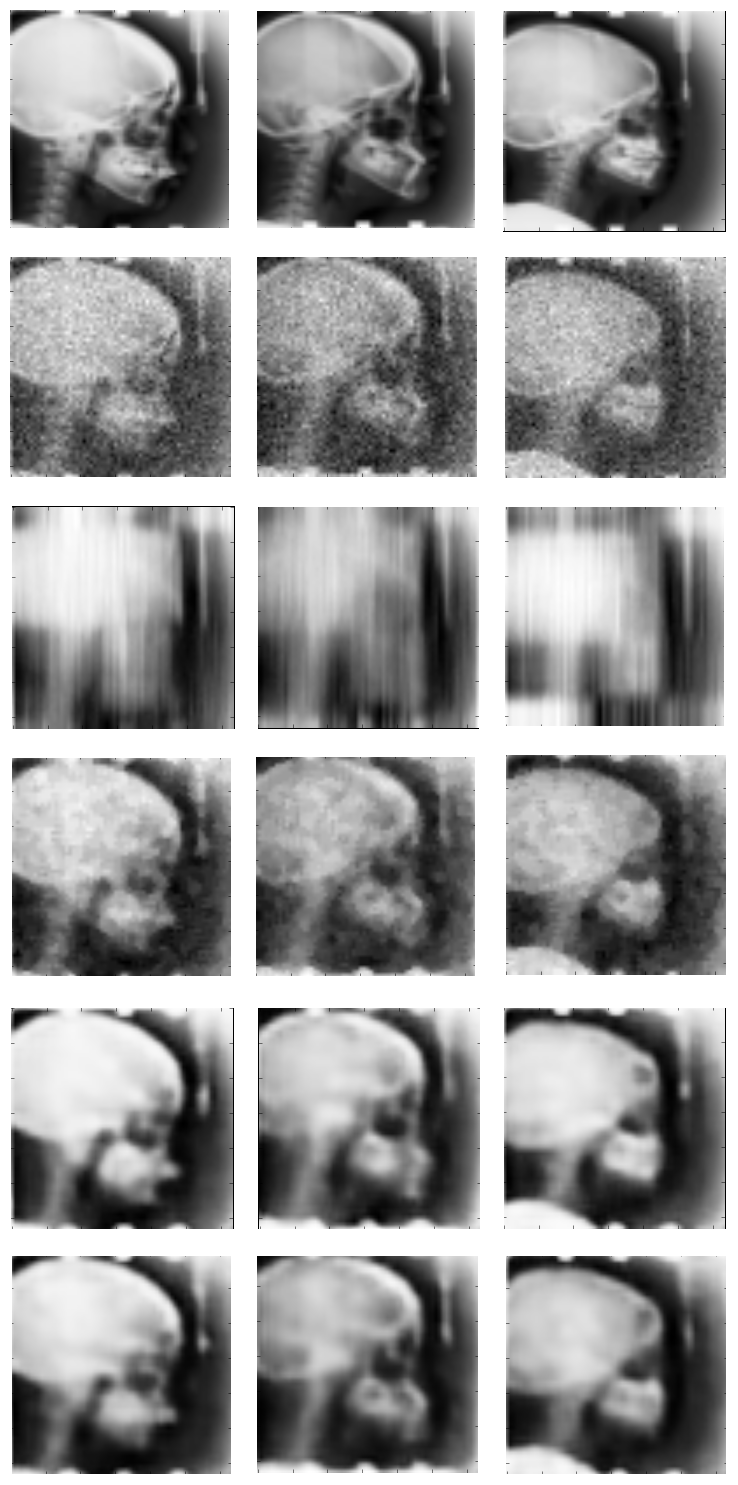}
  \caption{Denoising performance of CNN DAE on combined dataset, top row shows real images, second row is noisier version with minimal noise, third row is denoising result of NL means, fourth rows shows results of median filter, fifth row is results of using smaller dataset (300 training samples) with CNN DAE, sixth row is the results of CNN DAE on larger combined dataset.}
  \label{bigdatadenoise}
\end{figure}

Denoising results on three randomly chosen test images from combined dataset are shown in Fig. \ref{bigdatadenoise} and Table \ref{bigdatasets}. 

\begin{table}[h]
\centering
\caption{Comparing mean SSIM scores using different denoising filters}
\label{bigdatasets}
\begin{tabular}{|l|l|}
\hline
Image type    & SSIM \\ \hline
Noisy         & 0.63          \\ \hline
NL means      & 0.62          \\ \hline
Median filter & 0.80          \\ \hline
CNN DAE(a)    & 0.89         \\ \hline
CNN DAE(b)    & 0.90          \\ \hline
\end{tabular}
\vspace{5mm}
\\ {CNN DAE(a) is denoising performance using smaller dataset and CNN DAE(b) is denoising performance on same images using the combined dataset.}
\end{table}

Table \ref{bigdatasets} shows that CNN DAE performs better than NL means and median filter. Increasing sample size marginally enhanced the denoising performance.

\begin{figure}[h]
  \centering
  \includegraphics[scale=1]{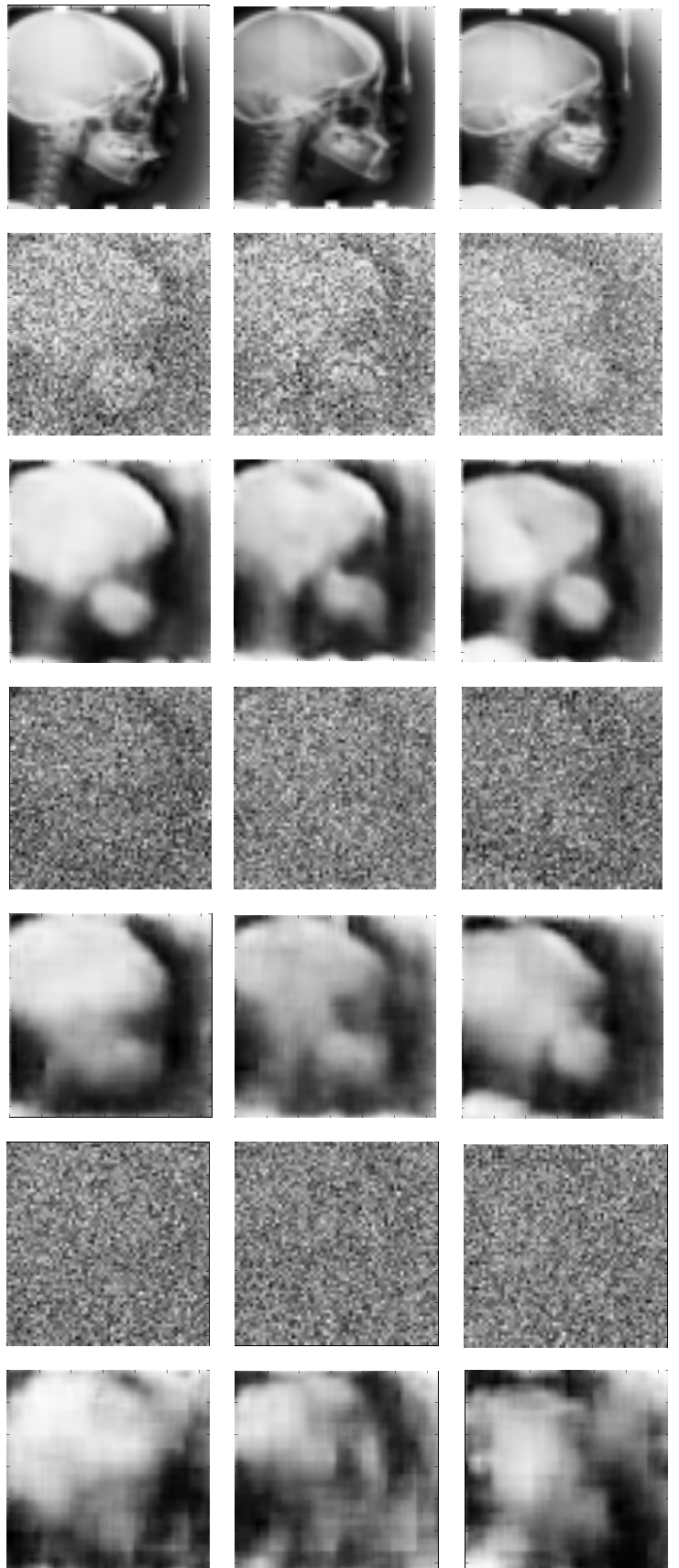}
  \caption{Denoising performance of CNN DAE on different Gaussian noise patterns. Top row shows original images, second row is noisy images with noise levels of $\mu=0, \sigma=1, p=0.5$, third row shows denoising results, fourth row shows corruption with $p=0.2, \sigma=5$, fifth row is denoised images using CNN DAE, sixth and seventh rows shows noisy and denoised images corrupted with $p=0.2, \sigma=10$.}
   \label{gausall}
\end{figure}

To test the limits of CNN DAEs denoising performance, we used rest of the noisy datasets with varying noise generative patterns and noise levels. Images with high corruption levels are barely visible to human eye, so denoising performance on those is of interest. Denoising results along with noisy and noiseless images on varying levels of Gaussian noise are shown in Fig. \ref{gausall}.

It can be seen that as noise level increases, this simple network has trouble reconstructing original signal. However, even when the image is not visible to human eye, this network is successful in partial generation of real images. Using a more complex deeper model, or by increasing number of training samples and number of epochs might help.

\begin{figure}[h]
  \centering
  \includegraphics[scale=1]{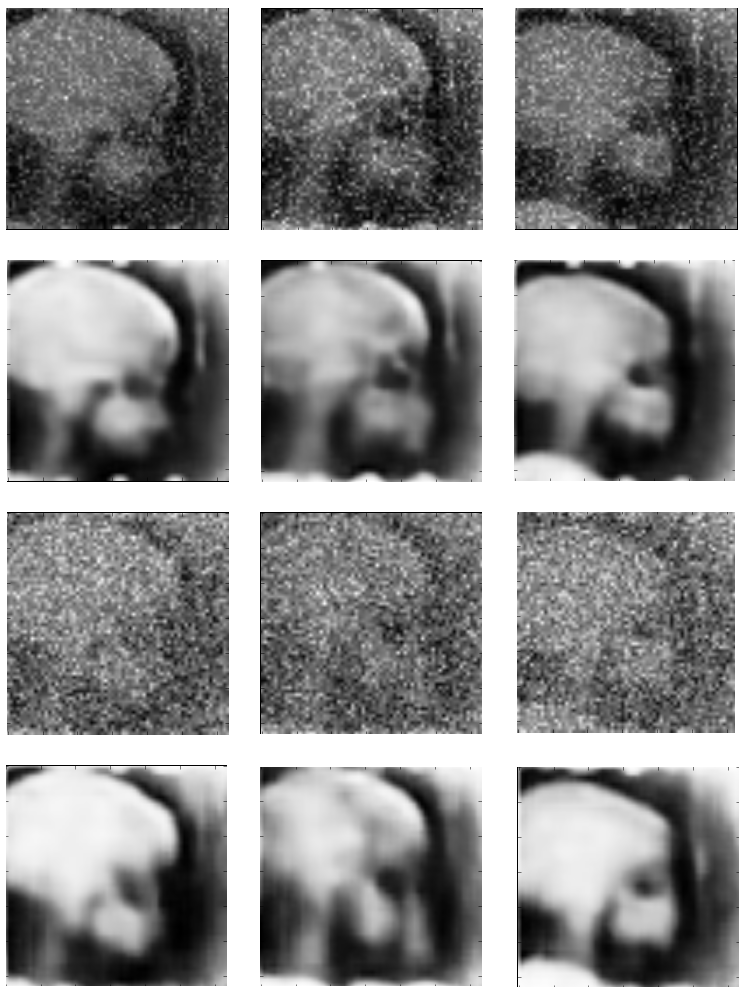}
  \caption{CNN DAE performance on Poisson corrupted images. Top row shows images corrupted with $p=0.2, \lambda=1$ with second row showing denoised results using CNN DAE. Third and fourth rows show noisy and denoised images corrupted with $p=0.2, \lambda=5$. }
  \label{poissonden}
\end{figure}

Performance of CNN DAE was tested on images corrupted using Poisson noise with  $p=0.2$, $\lambda=1$ and  $\lambda=5$. Denoising results are shown in Fig. \ref{poissonden}.

Table \ref{datasetsall} shows comparison of CNN DAE with median filter and NL means for denoising performance on varying noise levels and types. It is clear that CNN DAE outperforms both denoising methods by a wide margin, which increases as noise level increases.

\begin{table}[h]
\centering
\caption{Comparison using mean SSIM for different noise patterns and levels}
\label{datasetsall}
\begin{tabular}{|l|l|l|l|l|}
\hline
Image type    & $p=0.5$ & $sd=5$ & $sd=10$ & $Poisson, \lambda = 5$ \\ \hline
Noisy         & 0.10    & 0.03   & 0.01    & 0.33 \\ \hline
NL means      & 0.25    & 0.03   & 0.01    & 0.15 \\ \hline
Median filter & 0.28    & 0.11   & 0.03    & 0.17 \\ \hline
CNN DAE       & 0.70    & 0.55   & 0.39    & 0.85 \\ \hline
\end{tabular}
\vspace{5mm}
\\ {$p=0.5$ represents 50\% corrupted images with $\mu=0,\sigma=1$, $sd=5$ are images corrupted with $p=0.2, \mu=0, \sigma=5$, $sd=10$ are corrupted with $p=0.2, \mu=0, \sigma=10$ and $Poisson, \lambda=5$ are corrupted with a Poisson noise using $\lambda=5$}
\end{table}

\begin{figure}[h]
  \centering
  \includegraphics[scale=0.5]{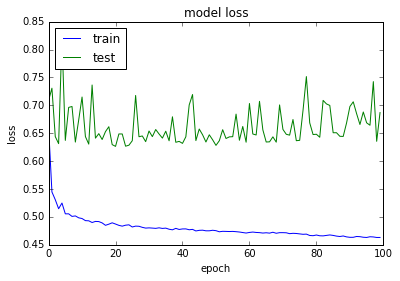}
  \caption{Model having trouble converging at higher noise levels, no decrease in validation errors can be seen with increasing number of epochs.}
  \label{troublecon}
\end{figure}

Also, as the noise level is increased the network has trouble converging. Fig. \ref{troublecon} shows the loss curves for Gaussian noise with $\mu=0, p=0.2, \sigma=10$. Even using 100 epochs, model has not converged.

\section{Conclusion}
We have shown that denoising autoencoder constructed using convolutional layers can be used for efficient denoising of medical images. In contrary to the belief, we have shown that good denoising performance can be achieved using small training datasets, training samples as few as 300 are enough for good performance. 

Our future work would focus on finding an optimal architecture for small sample denoising. We would like to investigate similar architectures on high resolution images and the use of other image denoising methods such as singular value decomposition (SVD) and median filters for image pre-processing before using CNN DAE, in hope of boosting denoising performance. It would also be of interest, if given only a few images can we combine them with other readily available images from datasets such as ImageNet \cite{Deng_2009} for better denoising performance by increasing training sample size.

\end{document}